\documentclass[letterpaper, 10 pt, conference]{ieeeconf}
\IEEEoverridecommandlockouts                              
\usepackage{cite}
\usepackage{amsmath,amssymb,amsfonts}
\usepackage{subfigure}
\usepackage{algorithmic}
\usepackage{graphicx}
\usepackage{subfigure}
\usepackage{textcomp}
\usepackage{xcolor}
\usepackage{color}
\usepackage{siunitx}

\title{\LARGE \bf Time-to-Collision-Aware Lane-Change Strategy Based on Potential Field and Cubic Polynomial for Autonomous Vehicles}

\author{Pengfei Lin$^{1}$,
    Ehsan Javanmardi$^{1}$,
    Ye Tao$^{1}$,
    Vishal Chauhan$^{1}$,
    Jin Nakazato$^{1}$, and
    Manabu Tsukada$^{1}$
\thanks{$^{1}$The authors are with the Department of Creative Informatics, Graduate School of Information Science and Technology, The University of Tokyo, Tokyo 113-8654, Japan 
    {\tt\small \{linpengfei0609,
    ejavanmardi, 
    vishalchauhan,
    jin-nakazato, 
    mtsukada\}@g.ecc.u-tokyo.ac.jp, 
    tydus@hongo.wide.ad.jp}}
}

\begin{document}

\maketitle

\begin{abstract}
Making safe and successful lane changes (LCs) is one of the many vitally important functions of autonomous vehicles (AVs) that are needed to ensure safe driving on expressways. Recently, the simplicity and real-time performance of the potential field (PF) method have been leveraged to design decision and planning modules for AVs. However, the LC trajectory planned by the PF method is usually lengthy and takes the ego vehicle laterally parallel and close to the obstacle vehicle, which creates a dangerous situation if the obstacle vehicle suddenly steers. To mitigate this risk, we propose a time-to-collision-aware LC (TTCA-LC) strategy based on the PF and cubic polynomial in which the TTC constraint is imposed in the optimized curve fitting. The proposed approach is evaluated using MATLAB/Simulink under high-speed conditions in a comparative driving scenario. The simulation results indicate that the TTCA-LC method performs better than the conventional PF-based LC (CPF-LC) method in generating shorter, safer, and smoother trajectories. The length of the LC trajectory is shortened by over 27.1\%, and the curvature is reduced by approximately 56.1\% compared with the CPF-LC method.
\end{abstract}


\section{Introduction}\label{intro}

Autonomous vehicles (AVs) are being designed to improve the quality and safety of people's travel experiences while raising the standards of intelligent transportation systems. They have the potential to reduce road accidents and improve overall social well-being, including increasing travel accessibility for disabled persons \cite{Ross2017-vl}, reducing traffic jams, providing intelligent collision avoidance, and completing a safe lane-change (LC) maneuvers \cite{Krasniqi2016-uv}. AVs are expected to do these things by automatically conducting fundamental driving tasks without human involvement \cite{Ignatious2022-jd} in natural traffic environments while ingesting real-time data provided by various onboard sensors and processed by deep-learning driving algorithms. 

Path Planning, a vital component of the autonomous driving system, is responsible for generating a collision-free path that considers several factors, including path length, curvature, etc. Therefore, many distinctive path-planning methods have been proposed in the past few years for AVs, such as the rapidly-exploring random tree star (RRT*) \cite{Noreen2016-dt}, state lattices \cite{Zhang2019-ry}, and the potential field (PF) \cite{Ji2017-fk}. Among those approaches, the PF is becoming widely used due to its fast real-time performance in finding the optimal path and simple structure with the given driving environment information. The PF method originated with electromagnetism, in which attractive and repulsive PFs are created for goal position and obstacles, respectively \cite{Khatib1986-dv}.

\begin{figure*}[t]
    \centering
    \includegraphics[width=\hsize]{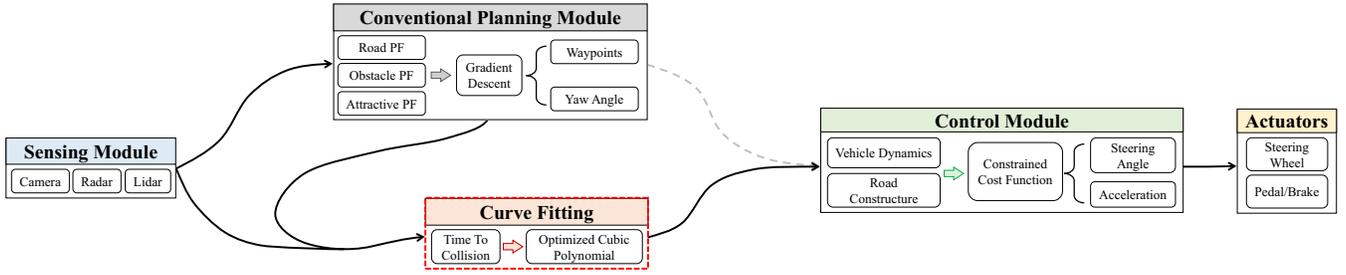}
    \caption{Proposed TTCA-LC strategy: Curve fitting computes the optimized cubic polynomial by considering the waypoints and yaw angle from the planning module as well as the TTC from the sensing module.}
    \label{system_scheme}
\end{figure*}

The PF method has recently been utilized in accomplishing the LC task which is one of the essential performance indexes for evaluating the autonomous driving system. Typically, the LC path is formulated by designing the specified geometry of the obstacle PF, such as the triangle \cite{Wolf2008-ye} and ellipse \cite{Ji2017-fk}. The path is generated along the edge of the obstacle PF using the gradient descent (GD) method, which conforms to the requirements of the lower control system from AVs due to its high real-time performance. However, the PF-based LC strategy for high-speed driving scenes usually generates a lengthy path due to the safety consideration of the PF. Besides, the finishing point of the PF-based LC is laterally parallel or close to the obstacle because the geometry design of the obstacle PF is based on the obstacle vehicle's center of gravity (c.g.). A potential collision risk exists if the adjacent obstacle suddenly steers without a pre-warning due to unexpected events. Hence, to improve the quality of the planned LC path, this study proposes a time-to-collision-aware LC strategy that uses the optimized cubic polynomial to process the path intended by the PF, as depicted in Fig. \ref{system_scheme}. The contributions of this paper are briefly summarized below:
\begin{itemize}
    \item An optimized cubic polynomial is proposed to process the waypoints from the PF that can generate an easy-to-follow trajectory.
    \item The TTC is used for constraint design in the optimization that can allow the AVs to complete the LC in advance and reserve sufficient space to prevent emergencies.
\end{itemize}

The remainder of this paper is organized as follows. Related works on PF-based LC methods are reviewed in Section II, and the proposed TTCA-LC method is described in Section III. Section IV presents the comparative simulation results, and conclusions are made in Section V.

\section{Related Works}\label{related_work}

This section reviews previous works on PF LC methods. In 2008, Wolf et al.~\cite{Wolf2008-ye} proposed a group of artificial potential functions to help AVs achieve collision avoidance, and a triangle-shaped potential field was established for the front obstacle vehicle that also played a role in guiding the LC behavior. Sharma et al.~\cite{Sharma2012-yw} presented a Lyapunov-based control scheme (LbCS) that designs attractive and repulsive potential functions to form a Lyapunov function and succeeded in making LC and merge maneuvers for car-like robots. A batch of lateral potentials was proposed by Kala et al.~\cite{Kala2012-kq} to handle driving scenarios in which speed lanes are absent; the lateral potentials govern only the vehicle steering. Galceran et al.~\cite{Galceran2015-oe} presented an integrated motion planning and control approach that uses PFs as driving corridors; this method enables a far smaller control effort based on the desired tracking tolerance. Subsequently, Ji et al. \cite{Ji2017-fk} and Rasekhipour et al.~\cite{Rasekhipour2017-yh} combined PF with model predictive control to guarantee that the planned trajectory conforms to the vehicle dynamics; a two-degree-of-freedom bicycle model was used to design the controller. Li et al. \cite{Li2017-dy} proposed a PF approach-based trajectory control strategy for electric AVs and formulated an innovative potential function to calculate the desired yaw angle. Lin et al. \cite{Lin2020-uc, Lin2022-jm} combined a PF with a clothoid curve, ensuring that the generated path is trackable. The clothoidal coefficients are constrained by a set of fixed parameters. Peng et al. \cite{Peng2022-sq} presented a hierarchical motion planning system for dynamic LC behavior, and the PF is modified to choose the optimal target lane while considering the speed differences between the ego and adjacent vehicles. Wu et al.~\cite{Wu2022-ae} proposed a new type of LC algorithm based on the PF that aims to generate human-like trajectories by considering environmental risks, driver focus shifts, and driver speed requirements.
\begin{figure}[t]
        \subfigure[3D view of road potential field]{
        \begin{minipage}{0.98\linewidth}
            \centering
            \includegraphics[width=\hsize]{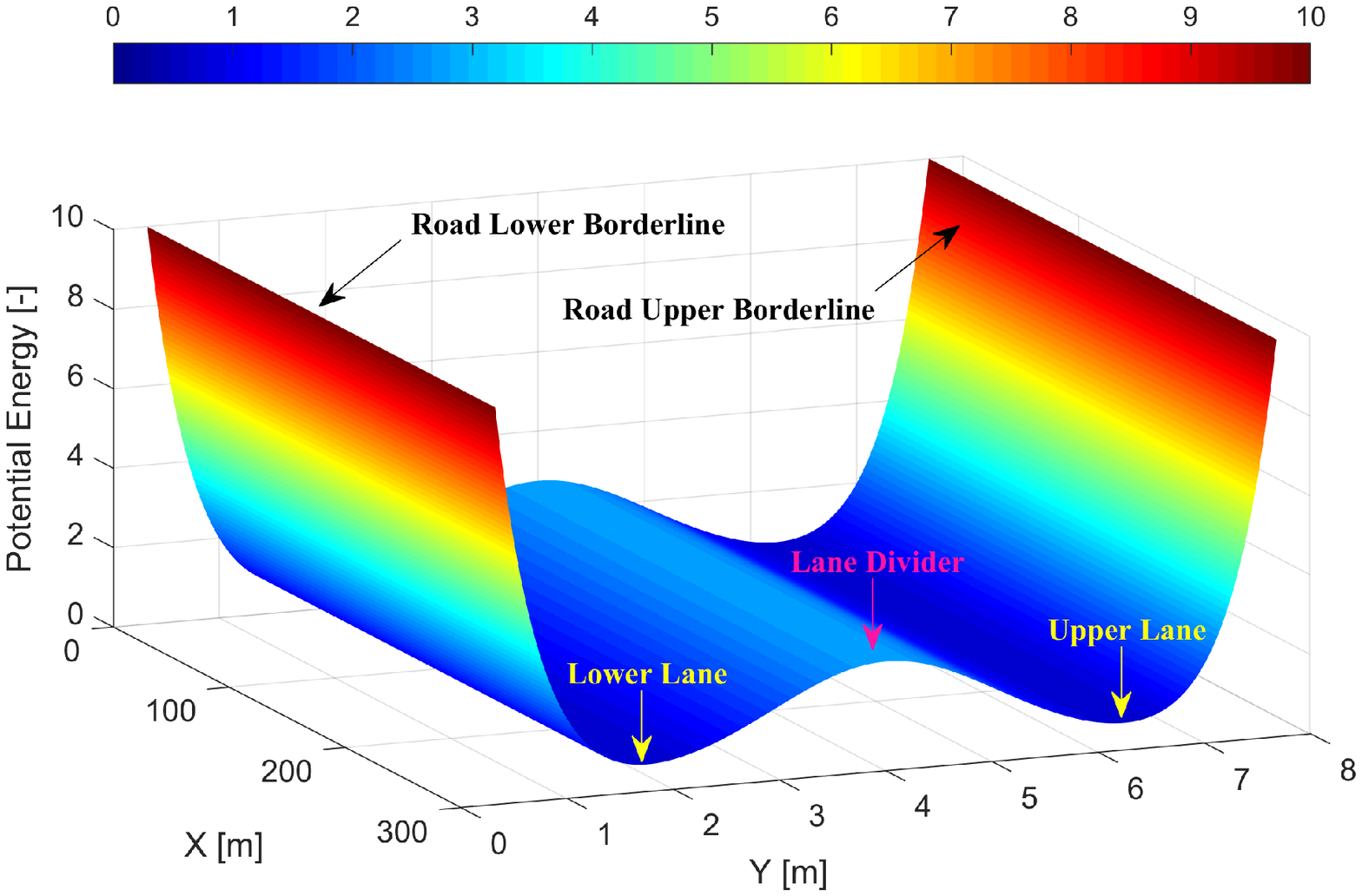}
            \label{road_pf_1}
        \end{minipage}
        }\\
        \subfigure[2D view of road potential field along the Y-Z axis]{
        \begin{minipage}{0.98\linewidth}
            \includegraphics[width=\hsize]{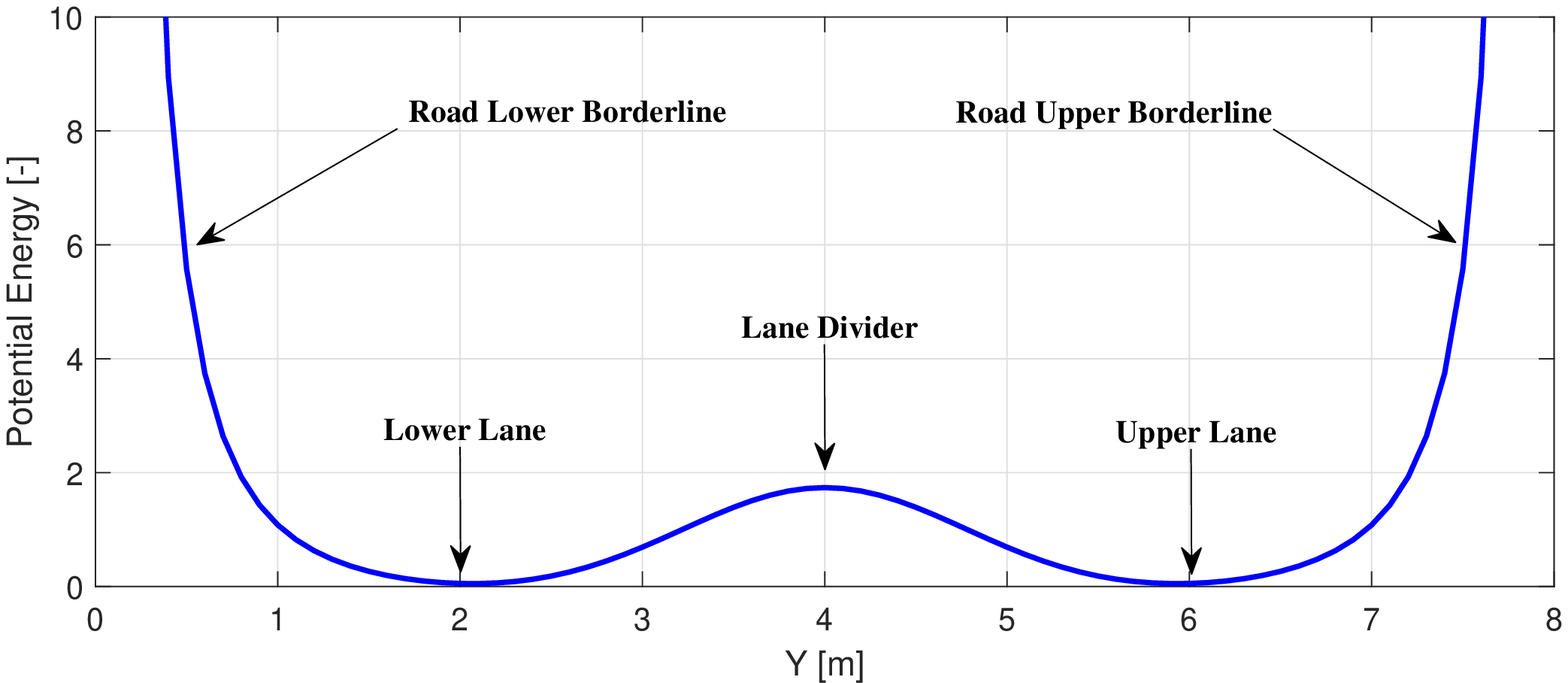}
            \label{road_pf_2}
        \end{minipage}
        }
\caption{Road potential field for a two-lane highway road.}
\end{figure}

The above PF-based LC methods generally depend on the geometry of the obstacle PF, in which the path is usually too long and laterally parallel or close to the obstacle vehicle's position. Furthermore, the curvature of the conventional PF-based LC (CPF-LC) path is sometimes sizeable, causing riding discomfort for passengers. The CPF-LC methods can also have a high potential collision risk if the obstacle vehicle suddenly steers toward the ego vehicle without an alert because of unexpected emergencies.

\section{PF-based Lane Change Strategy}\label{pf}

This section introduces the potential functions for establishing the PF and proposed cubic polynomial for processing waypoints.
\begin{figure*}[t]
    \centering
    \includegraphics[width=\hsize]{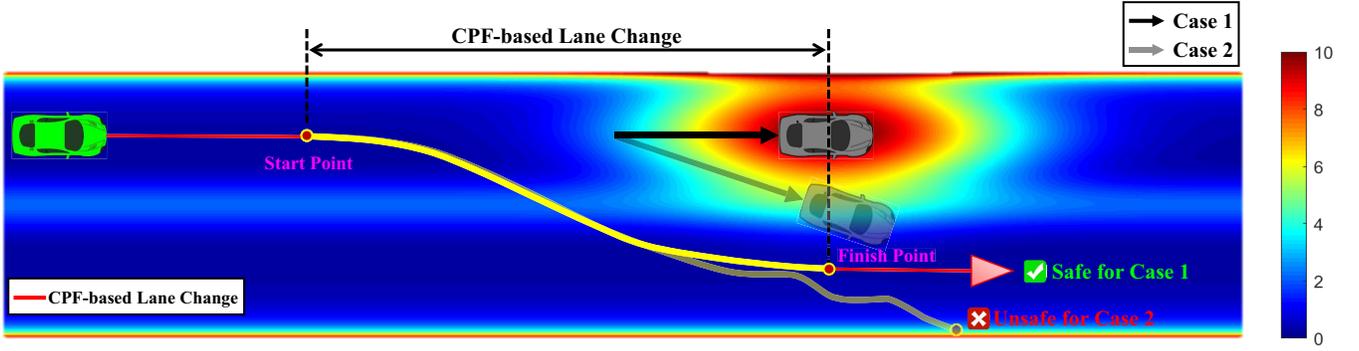}
    \caption{Conventional potential field-based lane-change strategy for different driving cases on the highway.}
    \label{cpf}
\end{figure*}

\subsection{Potential Functions}

As applied to road structures, the potential functions consist of three parts: attractive, road, and obstacle potentials.

\subsubsection{Attractive Potential} To drive a vehicle forward, the attractive potential function, $P_{a}$, is used to describe an attractive force:
\begin{equation}
    P_{a}=
    \frac{1}{2}\lambda \left(X-X_{tar}\right)^2,
    \label{attr_pf}
\end{equation}
where $\lambda$ denotes the slope scale, and $X$ and $X_{tar}$ are the longitudinal positions of the vehicle and target point, respectively. 

\subsubsection{Road Potential} A road generally comprises visible edges, lanes, and lane dividers. Hence, the road potential function, $P_{r}$, is formulated as follows:
\begin{equation}
    P_{r} =
    \frac{1}{2}\xi(\frac{1}{Y-Y_{l,u}-\frac{l_w}{2}})^{2} + A_{lane}\exp{-\frac{(Y-Y_{lane}^i)^2}{2\sigma^2}},
\end{equation}
where $\xi$ denotes the scaling coefficient, and $Y$ is the lateral position of the ego vehicle. $Y_{l,u}$ represents the lateral positions of both road edges. $l_w$ denotes the width of the vehicle. $A_{lane}$ is the amplitude of the lane divider's PF, and $Y_{lane}^i$ is the lateral position of $i^{th}$ lane divider. $\sigma$ is the rising/falling slope of the lane potential. The road potential is depicted in Figs. \ref{road_pf_1} and \ref{road_pf_2}.
\begin{figure}[t]
    \centering
    \includegraphics[width=0.9\hsize]{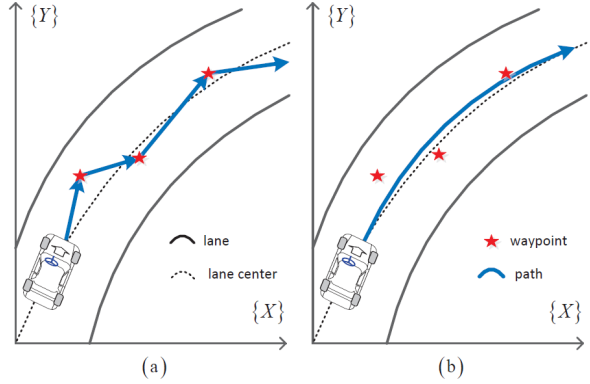}
    \caption{Waypoint tracking methods \cite{Jeon2015-vd}: (a) point-wise, (b) curve-fitting.}
    \label{wp_explain}
\end{figure}

\subsubsection{Obstacle Potential} An obstacle's PF is used to identify a boundary region that keeps the ego vehicle a safe distance away using graduated risk values. The obstacle’s PF plays a vital role in producing the LC trajectory, which typically flows along the edge of the PF boundary region. Therefore, the obstacle potential function, $P_o$, is applied~\cite{Lin2022-uc}:
\begin{align}
    P_o=
    A_{obs} \exp{[-\frac{C_1}{2}(\frac{(X-X_o)^2}{\sigma_x}+\frac{(Y-Y_o)^2}{\sigma_y}-C_2)]},
    \label{obs_pf}
\end{align}
where
\begin{align*}
    &C_1=1-\psi_o^2,\quad
    C_2=\frac{2\psi_o(X-X_o)(Y-Y_o)}{\sigma_x\sigma_y},\\
    &\sigma_x=D_{min}\sqrt{-\frac{1}{\ln{U}}},\quad
    \sigma_y=\sqrt{-\frac{(Y-Y_o)^2}{2\ln{\frac{U}{A_{obs}}}}},\\
    &D_{min}=\frac{Mv^2}{2a_b}-\frac{M_ov_o^2}{2a_{b,o}}+\frac{l_{fr}+l_{fr,o}}{2}.
\end{align*}
$X_o$ and $Y_o$ denote the longitudinal and lateral positions of the obstacle vehicle, respectively. $\psi_o$ is its heading angle, and $U$ is the minimum positive factor. $M$ and $M_o$ refer to the weights of the ego and obstacle vehicles, respectively. $v$ and $v_o$ denote the longitudinal speeds of the ego and obstacle vehicles, respectively. $a_b$ and $a_{b,o}$ are the maximum deceleration via braking of the ego and obstacle vehicles, respectively. $l_{fr}$ and $l_{fr,o}$ represent the wheelbases of the host and obstacle vehicles, respectively. Finally, we obtain the universal PF by summing Eqs. (\ref{attr_pf}) --(\ref{obs_pf}) and applying the gradient descent method \cite{Ruder2016-ve}. The LC trajectory is then obtained, as shown in Fig. ~\ref{cpf}.

\subsubsection{Optimized Cubic Polynomial} To produce a smooth and easy-to-follow trajectory, the cubic polynomial is frequently used for processing discrete waypoints, resulting in a smaller control effort for car-like robots \cite{Nagy2001-nv}. Similarly, the cubic polynomial is utilized for AVs when tracking Global Positioning System waypoints \cite{Jeon2015-vd}, as shown in Fig. \ref{wp_explain}. In this study, we propose including the TTC in the constraint design while solving the optimization. First, the standard cubic polynomial is formulated as
\begin{equation}
    f(x) = a_0 + a_1x + a_2x^2 + a_3x^3,
    \label{cubic}
\end{equation}
\begin{figure}[t]
    \centering
    \includegraphics[width=\hsize]{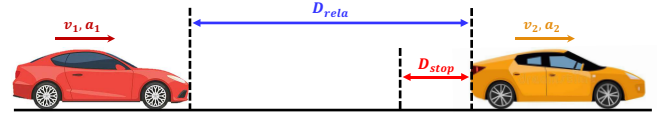}
    \caption{Two vehicles driving in the same direction: red vehicle (rear) drives at $(v_1,\;a_1)$; orange vehicle (front) drives at $(v_2,\;a_2)$.}
    \label{ttc}
\end{figure}
\begin{figure*}[t]
    \centering
    \includegraphics[width=\hsize]{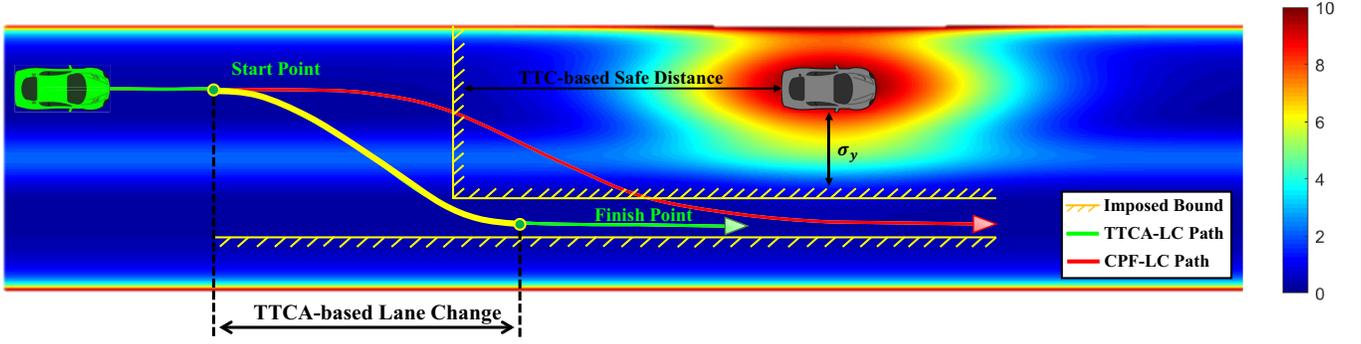}
    \caption{Proposed time-to-collision-aware lane-change strategy with the potential field: using the TTC for constraint design that allows the ego vehicle completes the LC in advance, reserving sufficient space for deceleration if the obstacle suddenly steers.}
    \label{ipf}
\end{figure*}
where $a_0,\;a_1,\;a_2,\;a_3$ are the parameters of the cubic polynomial, and $x$ represents the waypoints. Next, we reshape Eq. (\ref{cubic}) into a quadratic programming form for optimization:
\begin{equation}
    \mathbf{f} \, = \,
    \min_{\mathbf{a}}\frac{1}{2}\left(\textbf{X}\mathbf{a}-\textbf{Y}\right)^TW\left(\textbf{X}\mathbf{a}-\textbf{Y}\right),
\label{cost}
\end{equation}
where
\begin{align*}
&\textbf{\textrm{X}}=\begin{bmatrix} 1& x_1& {x_1}^2& {x_1}^3\\ 1& x_2& {x_2}^2& {x_2}^3\\ \vdots& \vdots& \vdots& \vdots\\ 1& x_N& {x_N}^2& {x_N}^3 \end{bmatrix},
W=\begin{bmatrix}
w_1 & 0  & \cdots   & 0  \\
0 & w_2  & \cdots   & 0  \\
\vdots & \vdots  & \ddots   & \vdots  \\
0 & 0  & \cdots\  & w_N \end{bmatrix},\\
&\mathbf{a}=\begin{bmatrix} a_0& a_1& a_2& a_3 \end{bmatrix}^T,~
\textbf{\textrm{Y}}=\begin{bmatrix} y_1& y_2& \cdots& y_N \end{bmatrix}^T.
\label{waypoint_matrix}
\end{align*}
$\mathbf{a}$ denotes the optimal cubic coefficient, and $(x_N,\;y_N)$ represents the $N^{th}$ waypoint from the PF. $w_N$ is the weight. Note that at least four waypoints are required to calculate $\mathbf{a}$. Next, we introduce the TTC, which represents the time required for the two vehicles to collide if they maintain the same speed driving along the same path \cite{Lee1976-ly}. As depicted in Fig. \ref{ttc}, we denote $D_{rela}$ as the relative distance between the vehicles, and $D_{stop}$ is the minimum safe stopping distance for the rear vehicle. We generally consider two collision cases when calculating TTC. The first is when the minimum $D_{rela}$ occurs after the front vehicle makes a complete stop, and the second is when the minimum $D_{rela}$ occurs before the front vehicle completes its stop. In the first case, we use the following formula:
\begin{equation}
    TTC=
    \begin{cases}
        \frac{D_{rela}-D_{stop}+\frac{v_2^2}{2|a_2|}}{v_1},& \text{\quad$a_1=0$}\\
        \frac{-v_1+\sqrt{v_1^2+2a_1(D_{rela}-D_{stop}+\frac{v_2^2}{2|a_2|})}}{|a_1|},& \text{\quad$a_1\neq0$},
    \end{cases}
\end{equation}
where $v_1$ and $a_1$ denote the speed and acceleration of the rear vehicle, respectively. $v_2$ and $a_2$ are the speed and acceleration of the vehicle in front, respectively. In the second case, the TTC is obtained as follows:
\begin{equation}
    TTC=
    \begin{cases}
        \frac{D_{rela}-D_{stop}}{v_{rela}},& \text{\quad$a_{rela}=0$}\\
        \frac{-v_{rela}+\sqrt{v_{rela}^2+2a_{rela}(D_{rela}-D_{stop})}}{a_{rela}},& \text{\quad$a_{rela}\neq0$},
    \end{cases}
\end{equation}
where
\begin{equation*}
    v_{rela}=|v_1-v_2|,\quad a_{rela}=|a_1-a_2|.
\end{equation*}
TTC is also an important indicator in the design of a forward collision warning system. Therefore, the standardization of TTC directly impacts traffic management. For example, an excessive TTC can lead to aggressive driving maneuvers, and an undersized TTC cannot ensure safety. The Mobileye Incorporation \cite{mobil-report} sets a 2.7-s TTC alert. Moreover, \cite{Ataelmanan2021-jf} found that approximately 93\% of observed drivers complete LC maneuvers in less than 5 s. More than half completed LCs around 3.5 s. Thus, we set the constraint based on the critical TTC value and the appropriate LC duration. As shown in Fig. \ref{ipf}, we impose the lower and upper bounds to regulate the cubic polynomial while fitting the waypoints from the PF. Hence, we can optimize Eq. (\ref{cost}) with bounded constraints as follows:
\begin{alignat}{2}
    \mathbf{f} \, = \, &
    \min_{\mathbf{a}}\frac{1}{2}\left(\textbf{X}\mathbf{a}-\textbf{Y}\right)^TW\left(\textbf{X}\mathbf{a}-\textbf{Y}\right)\\
    \mathrm{s.t.} \quad & \mathbf{X}\mathbf{a}\leq \sigma_{y},\quad \text{$if\;X_{start}\;\leq\;X\;\leq\;X_{end}$}\tag{8a}\\
    & \mathbf{X}\mathbf{a}\geq Y_{lb},\quad \text{$if\;X_{start}\;\leq\;X\;\leq\;X_{end}$,} \tag{8b}
\label{cost_with_cons}
\end{alignat}
where
\begin{align*}
    X_{start}&= X_{obs}-v(TTC+T_{LC}),\\
    X_{end}&=X_{obs}+\sigma_x,
\end{align*}
and $T_{LC}$ denotes the optimal LC duration. If the obtained TTC is less than the suggested threshold (2.7 s), the host vehicle must first brake to ensure longitudinal safety.
\begin{figure*}[ht]
    \centering
    \includegraphics[width=\hsize]{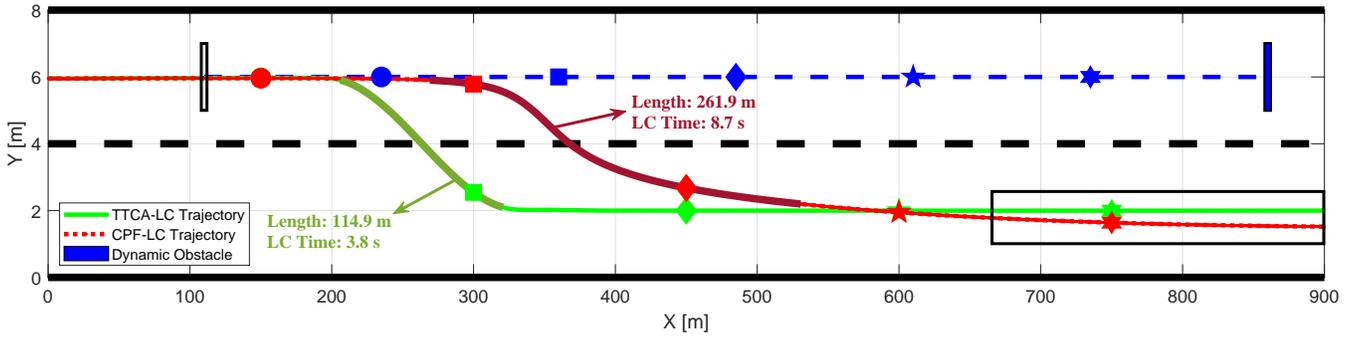}
    \caption{Trajectories of the conventional and optimized potential field planners under speeds of 108 km/h.}
    \label{path_2D}
\end{figure*}
\begin{figure}[t]
    \centering
    \includegraphics[width=\hsize]{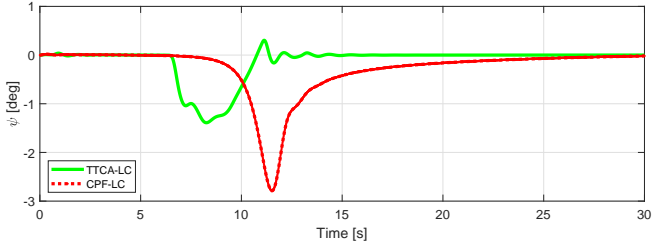}
    \caption{Yaw angle of the vehicle.}
    \label{psi}
\end{figure}
\begin{figure}[t]
    \centering
    \includegraphics[width=\hsize]{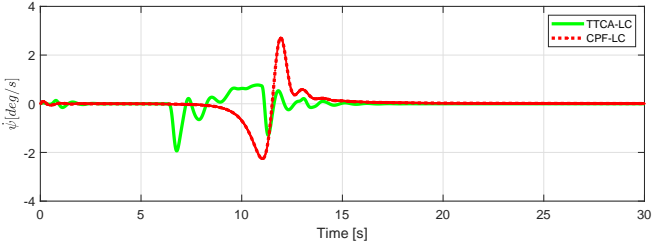}
    \caption{Yaw rates of the vehicles}
    \label{psi_dot}
\end{figure}
\begin{figure}[t]
    \centering
    \includegraphics[width=\hsize]{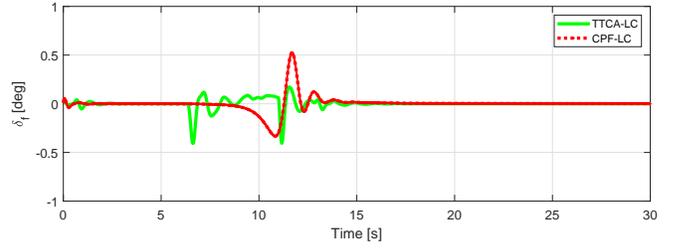}
    \caption{Front tire angles of the vehicles}
    \label{tire_steer}
\end{figure}
\begin{figure}[t]
    \centering
    \includegraphics[width=\hsize]{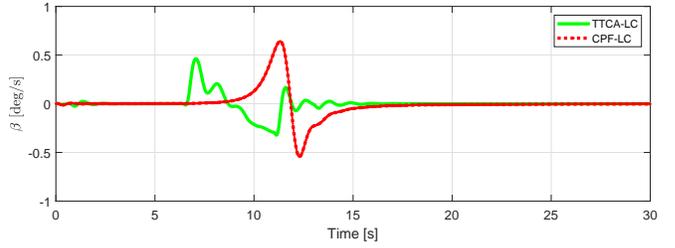}
    \caption{Sideslip angles of the vehicles}
    \label{sideslip}
\end{figure}
\begin{figure}[t]
    \centering
    \includegraphics[width=\hsize]{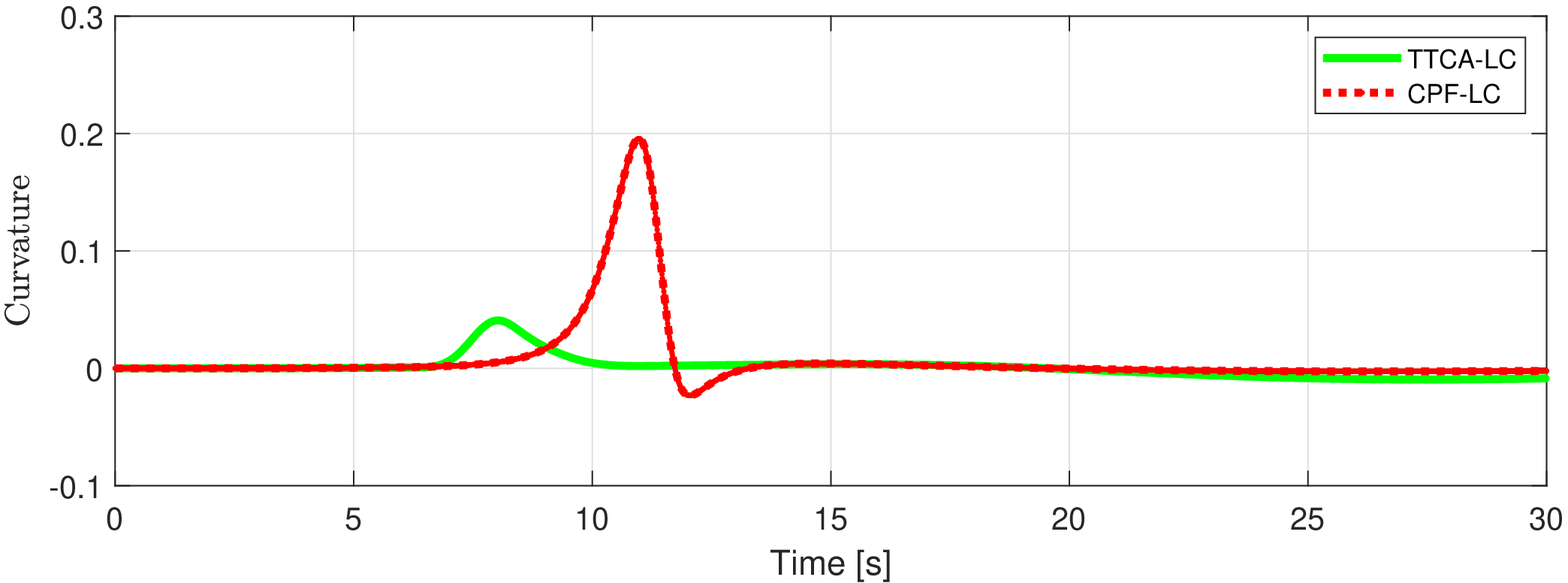}
    \caption{Curvature of the trajectories}
    \label{curva}
\end{figure}

\section{Simulation Results}\label{AA}

To compare our proposed TTCA-LC with the conventional PF, we set up a comparative scenario in which we simulated LC behaviors using two AV planning algorithms. The front vehicle was set as a road-bound moving obstacle whose initial parameters were $(120, 6)$ with an initial speed of 90 km/h. The ego vehicle spawned at (0, 6) in the same lane 120-m behind the obstacle vehicle. The initial speed of the ego vehicle was 108 km/h. According to \cite{Eckert2011-gt}, maintaining vehicle speed during the LC process is suitable for controlling the vehicle dynamics to ensure riding comfort. Therefore, we assumed that the speed of the ego vehicle was fixed during the LC. For comparison, we used the following two planners:
\begin{itemize}
    \item[i.] Conventional PF-based LC planner (denoted as CPF-LC) \cite{Wu2022-ae}
    \item[ii.] Time-to-collision-aware PF-based LC planner (denoted as TTCA-LC)
\end{itemize}

As depicted in Fig.~\ref{path_2D}, the blue trajectory shows the position of the dynamic obstacle along the x-y axis. The red and green lines show the trajectories of the ego vehicle using CPF-LC and TTCA-LC methods, respectively. As we can observe, the proposed TTCA-LC method exhibits better performance for different metrics. First, the TTCA-LC path length is reduced from 261.9 m to 114.9 m, which is 147.0 m shorter than that of the CPF-LC. Second, the LC start position of the TTCA-LC method is 63.0 m ahead of the CPF-LC method's position. Third, the LC time of the proposed method is 3.8 s which is closer to human drivers, while the LC time of CPF is 8.7 s. Finally, the trajectory of the CPF-LC deviates from the lane center after the LC, whereas the proposed TTCA-LC method keeps the ego vehicle at the lane center more accurately, as denoted by the black box in the figure.

Figs.~\ref{psi} and~\ref{psi_dot} show the yaw angles and rates of the vehicles. For the yaw angle, the CPF-LC method resulted in a relatively sharp turn of up to 3$^\circ$, at a rate of 2.7$^\circ$/s. By contrast, the TTCA-LC method cuts the maximum yaw angle by half, which is less than 1.5$^\circ$ throughout the entire process. Moreover, the yaw rate is also significantly reduced within 2$^\circ$/s, resulting in the direct flattening of the yaw angle. However, as illustrated in Fig. ~\ref{psi_dot}, the yaw rate of the green solid line forms an oscillation pattern at both the start and end of the LC. Both oscillations were mitigated in under 5 s, and the amplitude is relatively small so that passengers will not feel shaking. Our TTCA-LC method also significantly outperformed the CPF-LC methods in terms of yaw angle.

Fig.~\ref{tire_steer} shows the front tire steering angle during the LC maneuver. The CPF-LC and TTCA-LC methods showed relatively similar maximum front tire angles of approximately 0.52 and 0.41$^\circ$, respectively. According to \cite{Luciani2020-sy}, this range is within the limits of guaranteeing ride comfort among the LC duration.

Finally, Fig.~\ref{sideslip} shows the statistics for the sideslip angle. In the CPF-LC method, the ego vehicle performed sharp side slips ranging from -0.54$^\circ$ to 0.64$^\circ$. In contrast, the TTCA-LC method reduced the side slip to a maximum of ~0.45$^\circ$ with a slight oscillation. In Fig.~\ref{curva}, the curvature of the CPF-LC method reached 0.19$^\circ$ during the LC process. By contrast, the proposed TTCA-LC method reduced the curvature to 0.04$^\circ$, indicating that the path generated by the TTCA-LC method is smoother than that of the CPF-LC method.

From our microscopic analysis, the oscillations were mainly caused by the constraint imposed by Eq. (8); however, the duration amplitude of the oscillation was relatively short and small. Note that in all four angle charts, the TTCA-LC method produced slight oscillations at the beginning of the maneuvers because the initialization of the steady states of the cubic polynomial was not yet stabilized. Thus, they can be reduced by manually tuning the cubic parameters.

\section{Conclusion}

In this study, we have introduced a time-to-collision-aware lane-change (TTCA-LC) strategy that uses the PF with an optimized cubic polynomial. Furthermore, we detailed the formulation of the optimization function with reasonable constraints, in which the TTC was utilized to design the constraint for the safety considerations during the LC maneuver. The TTCA-LC method was validated in a comparative driving scenario via MATLAB/Simulink. The simulation results have indicated that the proposed method produces a safer, shorter (27.1\% shorten in length), and smoother (56.1\%  lower in curvature) LC path compared to the CPF-LC strategy, which reserves sufficient space for the braking maneuver if the adjacent obstacle drives abnormally.

In a future study, we will focus on multiple driving scenarios. For example, we plan to estimate the TTC from different heading angles instead of concentrating only on the longitudinal direction. Furthermore, the driving behaviors of additional surrounding obstacle vehicles will be defined more randomly, including sudden braking and acceleration behaviors. Furthermore, we will experiment with different speed scenarios to investigate the impact of the TTC-based constraint. 

\section*{Acknowledgment}

This work was supported by the commissioned research Grant number 01101 by the National Institute of Information and Communications Technology (NICT), Japan, and the Japan Society for the Promotion of Science (JSPS) KAKENHI (grant number: 21H03423), and partly sponsored by the China Scholarship Council (CSC) program (No.202208050036) and JSPS DC program (grant number: 23KJ0391).

\bibliographystyle{IEEEtran}
\bibliography{Reference.bib}

\end{document}